\documentclass{llncs}

\usepackage{helvet,times,courier}
\usepackage{graphicx}
\usepackage{amsmath}
\usepackage[usenames,dvipsnames]{color}
\usepackage{alltt}
\usepackage{verbatim}
\usepackage{mathnotation}
\usepackage{listings}

\definecolor{dark-gray}{gray}{0.25}

\makeatletter
\lst@Key{countblanklines}{true}[t]%
    {\lstKV@SetIf{#1}\lst@ifcountblanklines}

\lst@AddToHook{OnEmptyLine}{%
    \lst@ifnumberblanklines\else%
       \lst@ifcountblanklines\else%
         \advance\c@lstnumber-\@ne\relax%
       \fi%
    \fi}
\makeatother

\lstdefinelanguage{asp}{
    morecomment=[l]{\%},
    commentstyle=\itshape\color{dark-gray},
    captionpos=b, 
    numbers=left,
    numbersep=5pt,
    numberstyle=\tiny\color{dark-gray},
    numberblanklines=false,
    countblanklines=false,
    frame=bt, framexbottommargin=5pt, framextopmargin=5pt,
    aboveskip=5pt, belowskip=5pt,
    abovecaptionskip=10pt
}

\usepackage[draft]{fixme}
\fxsetup{
inline,
nomargin,
theme=color,
mode=multiuser
}
\FXRegisterAuthor{ks}{eks}{\color{blue}KS}
\FXRegisterAuthor{ar}{ers}{\color{magenta}AR}
\FXRegisterAuthor{gs}{egs}{\color{OliveGreen}GS}
\FXRegisterAuthor{fa}{efa}{\color{red}FA}

\DeclareMathOperator{\imp}{:\!-}
\DeclareMathOperator{\dnot}{\text{not}\;}

\newcommand{\id}[1]{\mathit{Id}_{\mathit{#1}}}

\renewcommand\texttt[1]{\mbox{\fontfamily{cmtt}\fontsize{9}{10}\selectfont#1}}

\title{OOASP: Connecting Object-oriented and Logic Programming\thanks{This research was funded by the Austrian Research Promotion Agency (grant number 840242) and Carinthian Science Fund (grant number KWF-3520/26767/38701)}}
\titlerunning{OOASP}  

\author{Andreas Falkner\inst{1} \and Anna Ryabokon\inst{2} \and Gottfried Schenner\inst{1} \and Kostyantyn Shchekotykhin\inst{2} }
\authorrunning{Falkner et al.} 
\institute{
Siemens AG \"Osterreich, Vienna, Austria\\
\email{firstname.\{middleinitial.\}lastname@siemens.com}
\and
Alpen-Adria-Universit{\"a}t Klagenfurt, Austria\\
\email{firstname.lastname@aau.at}
}

\begin{document}

\maketitle              

\setcounter{footnote}{0}
\renewcommand\thelstlisting{\arabic{lstlisting}}

\begin{abstract}
Most of contemporary software systems are implemented using an object-oriented approach. Modeling phases -- during which software engineers analyze requirements to the future system using some modeling language -- are an important part of the development process, since modeling errors are often hard to recognize and correct.

In this paper we present a framework which allows the integration of Answer Set Programming into the object-oriented software development process. OOASP supports reasoning about object-oriented software models and their instantiations. 
Preliminary results of the OOASP application in CSL Studio, which is a Siemens internal modeling environment for product configurators, show that it can be used as a lightweight approach to verify, create and transform instantiations of object models at runtime and to support the software development process during design and testing. 

\end{abstract}
%
\section{Introduction}\label{sec:intro}


Object-oriented programming languages is de facto a standard approach to software development. Many systems are modeled and implemented using it. In practice of Siemens the object-oriented approach is also used in many domains among which development of product configurators is one of the prominent examples.
A configurator is a software system that enables design of complex technical systems or services based on a predefined set of components. In modern configuration systems domain knowledge - comprising configuration requirements (product variability) and customer requirements - is expressed in terms of component types and relations between them. Each type is characterized by a set of attributes which specify functional and technical properties of real-world and abstract components of a configurable product. An attribute takes values from a predefined domain. Furthermore, components are related/connected to each other in various ways. 

Development of  object-oriented configurators is a challenging task due to several important issues such as acquisition of configuration knowledge from domain experts, modeling of this knowledge, model verification and maintenance. Different types of errors might occur, for example, due to the complexity of configuration models or procedural approach of object-oriented languages. Moreover, a variety of problems arises when configurable products or services have a long life-span and requirements are not stable, but change over time - for instance, if some components of a product are not produced any more or if a new functionality has to be added to a system. Some typical challenges occurring when a configuration is changed are discussed in~\cite{Falkner13}. Configuration technologies which address these tasks enable efficient production processes and thus can help reduce the overall production costs. 

Logic programming frameworks, such as Answer Set Programming (ASP), can  improve the speed and quality of object-oriented development. These frameworks provide expressive and easily understandable knowledge representation language allowing declarative encodings of complex problems. Equipped with powerful solving algorithms the logic programming frameworks showed their applicability in both product configuration as well as software development domains. For instance, important practical and theoretical aspects of formalizing real-world (re)configuration scenarios using a logic-based formalism are discussed in~\cite{Friedrich2011a}. The authors of~\cite{Falkner_testing} show how to support testing object-oriented and constraint-based configurators by automatically generating positive and negative test cases using ASP. A commercial ASP-based software for verification which makes the development of software easier and faster is suggested in~\cite{SchandaB12}.

In this paper we present an OOASP framework that uses a generic object-oriented configurator to encode its knowledge base and ASP for the computation of configurations. OOASP was implemented as an evaluation prototype for an extension to CSL Studio, an authoring environment for Configuration Specification Language (CSL)~\cite{DBLP:conf/gpce/DhunganaFH13}. It aims at the improvement of the software development process during design and testing. We illustrate the mapping from an object-oriented formalism (UML) to logical descriptions using a simplified real-world example from Siemens. Additionally, the paper provides different insights on (re)configuration tasks such as validation, completion and  reconciliation  of a configuration which can be accomplished by our system. 

The remainder of this paper is organized as follows. 
After a short ASP overview in Section~\ref{sec:asp}, we
describe in Section~\ref{sec:ooasp} how object-oriented knowledge bases can be specified using ASP within OOASP framework. In Section~\ref{sec:system} we introduce CSL Studio and discuss various product (re)configuration scenarios. Finally, in Section~\ref{sec:conc} we conclude and discuss the future work.

\section{Preliminaries}\label{sec:asp}

\paragraph{Answer set programming} (ASP) is an approach to declarative problem solving which has 
its roots in logic programming and deductive databases. It is a decidable
fragment of first-order logic interpreted under stable model semantics~\cite{Gelfond1988} and extended with default negation, aggregation, and optimization~\cite{Simons2002a}. ASP allows modeling of a variety of (combinatorial) search and optimization problems in a declarative way using model-based problem specification methodology (see e.g.~\cite{EiterIK09,Brewka2011} for details).

An ASP program $\Pi$ is a finite set of \emph{normal rules} of the form:
\begin{equation}
h \imp b_1, \dots, b_m, \dnot b_{m+1},\dots, \dnot b_n.
\label{eq:rule}
\end{equation} 
where 'not' denotes \emph{default negation}, $b_i$ ($0 \leq i$) and $h$ are atoms. An \emph{atom} is an expression of the form $p(\vec{t})$, where $p$ is a predicate and $\vec{t}$ is a vector of terms, i.e.\ constants, variables or uninterpreted function symbols~\cite{Gebser2011gr}. 
Extensions of ASP~\cite{Simons2002a} allow specific forms of atoms. Thus, a  \emph{cardinality constraint} is an atom of the form $l\{h_1, \dots, h_k\}u$, where $h_1,\dots, h_k$ are atoms and $l,u$ are non-negative integers. 
A \emph{literal} is either an atom $a$ or its negation $\dnot a$. 
In rule~(\ref{eq:rule}) the set of atoms $H(r)=\setof{h}$ 
is called \emph{head}, whereas the sets $B(r)^+=\{b_1,\dots,b_m\}$ and $B(r)^-=\{b_{m+1},\dots,b_n\}$ are \emph{positive} body and \emph{negative} body, respectively. A \emph{fact} is a rule $r$ with $B(r)^+ \cup B(r)^-=\emptyset$; an \emph{integrity constraint} is a rule $r$ with $H(r)=\emptyset$; and a \emph{choice} rule has a cardinality constraint as the head $h$.
A literal, rule or program is \emph{ground}, if it is variable-free. A non-ground program $\Pi$ can be grounded by substituting variables with constants appearing in $\Pi$. 

Semantics of a ground normal program $\Pi$ is defined in terms of Gelfond-Lifschitz reduct. Let $A(\Pi)$ be a set of atoms appearing in $\Pi$, then $I\subseteq A(\Pi)$ is an \emph{interpretation}. A Gelfond-Lifschitz reduct~\cite{Gelfond1988} of a program $\Pi$ wrt.\ an interpretation $I$ is defined as $\Pi^I=\setof{H(r)\leftarrow B^+(r)\; |\; r \in \Pi, I \cap B^-(r)=\emptyset}$. An interpretation $I$ is an \emph{answer set} of $\Pi$, if $I$ is a minimal model of $\Pi^I$. The semantics of a ground program $\Pi$ with cardinality contraint atoms is defined similarly, since each rule with such atoms can be translated into a set of normal rules~\cite{Simons2002a}. Informally, semantics of a cardinality constraint requires at least $l$ and at most $u$ atoms $h_i$ to be in an answer set.

Moreover, ASP allows finding of preferred answer sets. The preferences are defined by \emph{weak constraints} -- a specific type of integrity constraints that can be violated. Each violation is penalized by a weight associated with a constraint. Given a program with weak constraints an ASP solver returns an answer set minimizing the sum of penalties.


\section{OOASP framework}\label{sec:ooasp}

The development of an object-oriented software is a complex and error-prone activity that requires careful modeling of an underlying problem. Siemens experience in the development of industrial applications shows that quite often incorrect models are responsible for faults in software artifacts that are hard to identify and debug. In this section we present the OOASP approach which allows to analyze object-oriented software models and their instances by means of ASP. In particular, we consider those models that can be described by a modeling language corresponding to a UML class diagram~\cite{RumbaughJacobsonBooch05}. The latter is a language allowing a software developer to specify an object model and additional constraints that each valid instantiation of an object model must satisfy. 

In order to reason about a software model, OOASP framework uses a \emph{meta-program\-ming} approach~\cite{sterling1994art} which was successfully applied in a similar way, for instance, to debugging of ASP programs~\cite{Gebser2008b,Oetsch2010a}. In our meta-programming approach an ASP program over a meta-language manipulates an ASP program describing a software model in terms of the Domain Description Language (DDL). 
In case of OOASP, all concepts of one or multiple software models as well as their instantiations are represented in OOASP-DDL as a set of rules of the form~(\ref{eq:rule}). Then, a meta-program, designed to accomplish a specific reasoning task, is applied to a program in OOASP-DDL. In a standard implementation of OOASP we provide meta-programs accomplishing the following tasks\footnote{OOASP code and encodings are available upon request from the first author.}: 
\begin{description}
\item[Validation]
Given an OOASP-DDL program describing an object-oriented model and its instantiation, a validation meta-program verifies whether all integrity and domain-specific constraints hold. The integrity constraints encode model requirements to relations between objects of an instantiation and are derived from the given model automatically. The domain-specific constraints ensure that some specific requirements to an instantiation of a model are satisfied. They can either be directly specified in the meta-program or imported from other languages. For instance, one could import domain-specific constraints defined in Object Constraint Language\footnote{OCL specification is available from http://www.omg.org/spec/OCL/2.4/PDF/} (OCL), for which transformations to SAT~\cite{Soeken1871248} and constraints programming~\cite{4566993} exist. 

\item[Completion]  
Given an OOASP-DDL program describing an object-oriented model and its (partial) instantiation, the completion task is to find an extension of the instantiation that satisfies all constraints or to show that such extension does not exist. 
The latter may occur due to two main reasons: (i) the object-oriented model or the given (partial) instantiation are inconsistent and do not have a completion -- an empty instantiation can be seen as a special case for the completion; and (ii) the extension of the given instantiation requires the creation of a number of objects that exceeds the given upper bounds for object instances. 

\item[Reconciliation] 
Given an OOASP-DDL program describing a legacy instantiation of an outdated object-oriented model, a new up-to-date model and a set of transformation rules, the goal of the reconciliation is to find a possibly preferred set of changes required to transform the legacy instantiation to a valid instantiation of the new model. 
The preferences in OOASP can be defined with domain-specific costs that assess the costs of required changes such as creation, reuse or disposal (deletion) of object instances.
\end{description}
If advanced features such as multiple inheritance, symmetry breaking, etc., are required, the default ASP encodings of reasoning tasks, outlined in this paper, must be replaced with alternative encodings, whereas the OOASP-DDL program remains the same. 

\subsection{OOASP Domain Description Language} 

OOASP-DDL allows a software developer to define all standard concepts of object-oriented models such as classes, attributes and associations. 
Each concept of the model is translated to a corresponding OOASP-DDL atom, where each term $\id{*}$ is an identifier of a model, class, attribute, etc. In OOASP identifiers of models are globally unique, whereas all other identifiers are unique within a model.
In the current version OOASP-DDL supports the definitions presented in Table~\ref{tab:mod}.
These definitions are sufficient to describe a subset of the object-oriented model of programming languages such as C++, Java, etc. Many features that can additionally be found in object-oriented models, e.g.\ initial values, constants, multi-valued attributes, ordered associations, etc., are currently not supported by the framework. This is because our main purpose was to provide a lightweight approach that, however, is able to capture most of the features commonly used in practice.
The definition of an instantiation of an object-oriented model is done using OOASP-DDL in a similar way as the definition of the model. In particular, our language allows the definitions shown in Table~\ref{tab:inst}.

\begin{table}[tb]
	\centering
		\begin{tabular}{p{.43\textwidth}p{.57\textwidth}}
        $ooasp\_class(\id{M},\id{C})$ & a class $C$ is defined in a model $M$ \\
        $ooasp\_subclass(\id{M},\id{C},\id{SC})$ &  defines a subclass relation between a class $C$ and a super class $\mathit{CS}$ in a model $M$ \\
        $ooasp\_assoc(\id{M},\id{A}, \id{C_1},\mathit{Min}_{C_1},$         
        $\phantom\qquad\mathit{Max}_{C_1}, \id{C_2}, \mathit{Min}_{C_2}, \mathit{Max}_{C_2})$ & defines an association relation $A$  between classes $C_1$ and $C_2$ with the given cardinalities, e.g. for every instance of the class $C_1$ at least $\mathit{Min}_{C_2}$ and at most $\mathit{Max}_{C_2}$ instances of the class $C_2$ must be associated \\
                
      $ooasp\_attribute(\id{M}, \id{C}, \id{AT},$       
      $\phantom\qquad\{\text{\emph{``string'',``integer'',``boolean''}}\})$ & an attribute $\mathit{AT}$ of a class $C$ is defined to have one of the three possible types \\

      $ooasp\_attribute\_minInclusive(\id{M},$       
      $\phantom\qquad\id{C}, \id{AT}, \mathit{MinV})$ & provides an optional minimum value $\mathit{MinV}$ for an integer attribute $\mathit{AT}$ \\

      $ooasp\_attribute\_maxInclusive(\id{M},$       
      $\phantom\qquad\id{C}, \id{AT}, \mathit{MaxV})$ & provides an optional maximum $\mathit{MaxV}$ for an integer attribute $\mathit{AT}$ \\
      
      $ooasp\_attribute\_enum(\id{M},$       
      $\phantom\qquad\id{C}, \id{AT}, \mathit{Val})$ & defines a possible value $\mathit{Val}$ for a string attribute $\mathit{AT}$
		\end{tabular}    
    \caption{OOASP-DDL definitions for the encoding of models}\label{tab:mod}
\end{table}

\begin{table}[tb]
	\centering
		\begin{tabular}{p{.43\textwidth}p{.57\textwidth}}
        $ooasp\_instantiation(\id{M},\id{I})$ & defines an instantiation $I$ of a model $M$ \\
        $ooasp\_isa(\id{I},\id{C},\id{O})$ &    declares that an object $O$ is an instance of the class $C$ \\
        $ooasp\_associated(\id{I}, \id{A}, $ 
        $\phantom\qquad \id{O_1}, \id{O_2})$ & objects $O_1$ and $O_2$ are associated by the association relation $A$ \\
        $ooasp\_attribute\_value(\id{I},\id{AT},$
        $\phantom\qquad \id{O}, \mathit{Val})$ & assigns a value $\mathit{Val}$ to an attribute $\mathit{AT}$ of an object $O$       
		\end{tabular}       
    \caption{OOASP-DDL definitions for the encoding of instantiations}\label{tab:inst}
\end{table}

Note that, OOASP-DDL is designed in a way to allow the definition of multiple models and their instantiation in one ASP program. This provides the necessary support for reconciliation and similar reasoning tasks that are applied to many models and/or their instantiations at once. 

\subsection{Definition of constraints} 
Constraints allow a software developer to ensure that models and their instantiations are valid. In OOASP we support two types of constraints: integrity constraints and domain-specific constraints. 
The latter are used to verify some specific properties of a model and/or its instantiations. The definition of domain-specific constraints can be done by a developer directly in OOASP-DDL or by importing them from the input model, e.g.\ OCL constraints from a UML model. 
The integrity constraints, however, are included in the default OOASP implementation and capture the requirements of the input object-oriented model such as cardinality restrictions, typing, etc. For instance, in order to ensure that a minimal cardinality requirement of an association relation holds in a given instantiation, OOASP framework comprises the following rule\footnote{In our examples we use the gringo~\cite{Gebser2011gr} dialect of ASP that also allows usage of uninterpreted function symbols such as \emph{mincardviolated}.}:
\begin{lstlisting} 
ooasp_cv(I,mincardviolated(O1,A)) :-  
    {ooasp_associated(I,A,O1,O2): ooasp_isa(I,C2,O2)} C2MIN-1, C2MIN>0,    
    ooasp_assoc(M,A,C1,C1MIN,C1MAX,C2,C2MIN,C2MAX), 
    ooasp_instantiation(M,I), 
    ooasp_isa(I,C1,O1).
\end{lstlisting}
The presence of an atom over \emph{ooasp\_cv} predicate in an answer set of an OOASP program indicates that a corresponding integrity constraint is violated by the given instantiation. In the sample rule above, the error atom is derived whenever less objects of type $C_2$ are associated with object $O_1$ than required by the cardinality restriction of the association. 

\section{System description}\label{sec:system}

OOASP was implemented as a potential extension to any object-oriented modeling environment and its practicability was evaluated together with CSL Studio~\cite{DBLP:conf/gpce/DhunganaFH13}. The latter is a Siemens internal tool for the design of product configurators as Generative Constraints Satisfaction Problems (GCSPs)~\cite{DBLP:journals/expert/FleischanderlFHSS98,Stumptner1998b}. 
CSL (Configuration Specification Language) is a formal modeling language based
on a standard object-oriented meta-model similar to Ecore\footnote{Eclipse Modeling Framework https://www.eclipse.org/modeling/emf/} or MOF\footnote{MetaObject Facility http://www.omg.org/mof/}. 
It provides all state-of-the-art features such as packages, interfaces, enumerations, classes with attributes of various types, associations between classes, inheritance and aggregation relations. In addition, it offers reasoning methods such as rules and constraints which are not covered in this work. The reason is that they are not (yet) translated into OOASP domain-specific constraints. 

\begin{figure}[bt]
\centerline{\includegraphics[width=\textwidth]{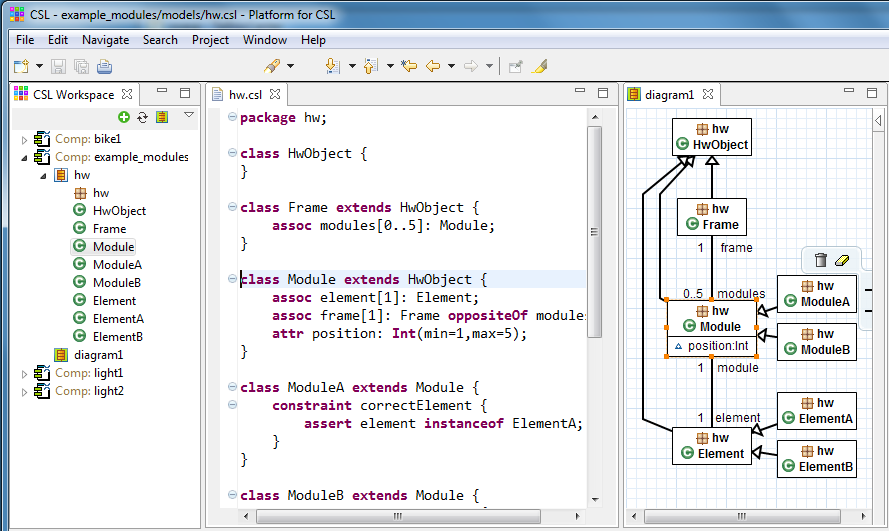}}
\caption{CSL screenshot for the Modules example}
\label{fig:cslexample}
\end{figure}

A screenshot of CSL Studio, presented in Fig.~\ref{fig:cslexample}, shows an example of a simple hardware configuration problem. A configuration problem corresponds to a composition activity in which a desired configurable product is assembled by relating individual components of predefined types. The components and relations between them are usually subject to constraints expressing their possible combinations allowed by the system's design.
The types of the components, relations between them as well as additional constraints on sets of related components constitute \emph{configuration requirements}. Many of those constraints can be expressed in an object-oriented model as cardinalities of association and aggregation relations.

The sample model shown in Fig.~\ref{fig:cslexample} describes a product configuration problem as a UML class diagram. In this problem the hardware product consists of a number of \emph{Frames}. Each frame contains up to five \emph{Modules} of types \emph{ModuleA} or \emph{ModuleB}, where each module occupies exactly one of the 5 positions in a frame. Moreover, each module has exactly one \emph{Element} assigned to it. All elements are of one of two types \emph{ElementA} or \emph{ElementB}. The corresponding OOASP-DDL encoding for this example is automatically generated by CSL Studio. A part of the encoding excluding integrity constraints is shown in Listing~\ref{lst:example}.

\begin{figure}[bt]%
\begin{lstlisting} [caption=OOASP-DDL encoding of the Modules example shown in Fig.~\ref{fig:cslexample},label=lst:example]
% modules example kb "v1"
% classes
ooasp_class("v1","HwObject").
ooasp_class("v1","Frame").
ooasp_class("v1","Module").  
ooasp_class("v1","ModuleA").  ooasp_class("v1","ModuleB").
ooasp_class("v1","Element"). 
ooasp_class("v1","ElementA"). ooasp_class("v1","ElementB").

% class inheritance
ooasp_subclass("v1","Frame","HwObject"). 
ooasp_subclass("v1","Module","HwObject").
ooasp_subclass("v1","Element","HwObject").
ooasp_subclass("v1","ElementA","Element"). 
ooasp_subclass("v1","ElementB","Element").
ooasp_subclass("v1","ModuleA","Module").
ooasp_subclass("v1","ModuleB","Module").

% attributes and associations
% class Frame
ooasp_assoc("v1","Frame_modules","Frame",1,1,"Module",0,5).

% class Module
ooasp_attribute("v1","Module","position","integer").
ooasp_attribute_minInclusive("v1","Module","position",1).
ooasp_attribute_maxInclusive("v1","Module","position",5).

% class Element
ooasp_assoc("v1","Element_module","Element",1,1,"Module",1,1).
\end{lstlisting}
\end{figure}

Additionally to the integrity constraints, implied by the cardinalities of associations shown on the UML diagram, there are the following domain-specific constraints:
\begin{itemize}
\item Elements of type \emph{ElementA} require a module of type \emph{ModuleA}
\item Elements of type \emph{ElementB} require a module of type \emph{ModuleB}
\item Modules must occupy different positions in a frame
\end{itemize}
These constraints can easily be implemented in OOASP. For instance, the first and the third can be formulated as shown in Listing~\ref{lst:const}.
\begin{figure}[tb]%
\begin{lstlisting}[caption=Sample domain-specific constraints in OOASP,label=lst:const]    
ooasp_cv(I,module_element_violated(M1,E1)) :-    
    ooasp_instantiation(M,I), 
    ooasp_associated(I,"Element_module",M1,E1),
    ooasp_isa(I,"ElementA",E1), 
    not ooasp_isa(I,"ModuleA",M1).

ooasp_cv(I,alldiffviolated(M1,M2,F)) :-    
    ooasp_instantiation(M,I), 
    ooasp_isa(I,"Module",M1), 
    ooasp_isa(I,"Module",M2),    
    ooasp_attribute_value(I,"position",M1,P), 
    ooasp_attribute_value(I,"position",M2,P),
    ooasp_associated(I,"Frame_modules",F,M1),
    ooasp_associated(I,"Frame_modules",F,M2),
    M1 != M2.   
\end{lstlisting}
\end{figure}

\begin{figure}[tb]
	\centering
		\includegraphics[width=0.8\textwidth]{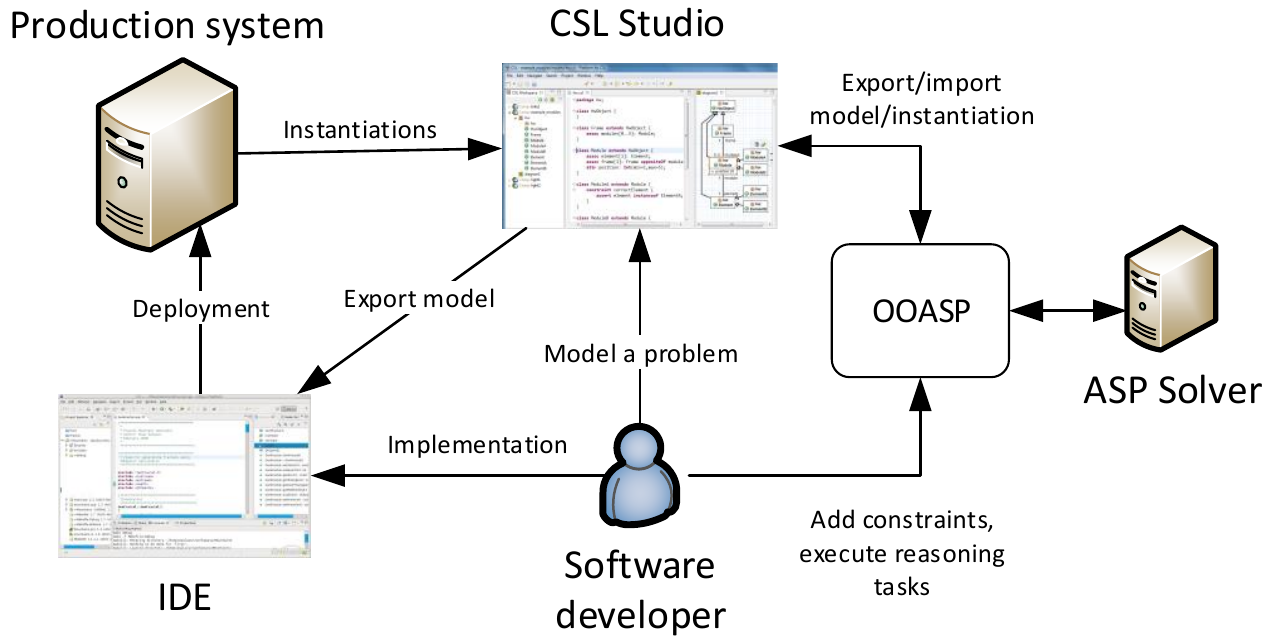}
	\caption{Integration of OOASP in development of product configurators}
	\label{fig:ooasp-workflow}
\end{figure}

A typical workflow of the product configurator development process in CSL Studio and OOASP is depicted in Fig.~\ref{fig:ooasp-workflow}. The development starts with a creation of an initial configuration model in CSL. Then, the model can be exported to OOASP and extended by the definition of domain-specific constraints.
Finally, the consistency of the developed model can be verified by execution of different reasoning tasks. For instance, the existence of model instantiations can be checked by running a completion task with an empty instantiation. The validation task can be used to test whether some of the known valid product configurations are instantiations of the model. Moreover, OOASP can be used during the implementation phase. Thus, CSL Studio allows a software developer to export a created model to a preferred object-oriented language as a set of classes. These generated classes must then be extended with the implementation of domain-specific constraints as well as additional methods and fields required for correct functionality of the software. In order to ensure that the software is implemented correctly, the software developer can export a (partial) instantiation generated by an object-oriented program to OOASP. In this case the completion reasoning task allows to test whether the obtained partial solution can be extended to a complete one, e.g.\ by creating missing modules for the elements as well as by adding missing frames and assigning the modules to them. 
In addition, if the software developer (tester) manipulates a completed configuration, for instance, by adding or removing elements, the configurator can restore consistency through reconciliation. The latter finds a set of changes that keep as much of the existing structure of the configured system as possible. In the following subsections we describe some use cases exemplifying OOASP applications during the development of configurators.

\subsection{Validation of a configuration}

The implementation of an object-oriented software requires continuous testing in order to identify and resolve faults early. The validation reasoning task provided by OOASP allows a software developer to verify whether an instantiation generated by the object-oriented code is consistent.
Especially, the validation is important in the context of CSL Studio or similar systems while testing domain-specific constraints. Thus, in CSL Studio an instantiation of the object model provided by the software developer is automatically exported to OOASP and the validation meta-program is executed. The obtained answer set is then used to highlight the parts of the instantiation that violate requirements to a valid configuration. Using this information, the developer can identify the faults in the software in a shorter period of time. 

For instance, assume a software developer implemented a model designed in CSL Studio and the resulting program outputs an instantiation \texttt{c2} comprising only one element of type \texttt{ElementA}. CSL Studio forwards this instantiation to OOASP which translates it to the OOASP-DDL program:
 \begin{center}
 \texttt{ooasp\_isa("c2","ElementA",10)}.
 \end{center}
For this input, execution of the validation task returns an answer set comprising:
\begin{center}
\texttt{ooasp\_cv("c2",mincardviolated(10,"Element\_module"))} 
\end{center}
This atom indicates that cardinality restrictions of the association between Element and Module classes are violated. The reason is that for the object with identifier \texttt{10} there is no corresponding object of the Module type.

Note that in the current OOASP prototype domain-specific constraints must be coded by a software developer manually and are not generated from the CSL (constraint language). However, this behavior was found to be advantageous in practice, since it provides a mechanism for the diverse redundancy~\cite{Falkner_testing}. The latter refers to the engineering principle that suggests application of two or more systems. These systems are built using different algorithms, design methodology, etc., to perform the same task. The main benefit of the diverse redundancy is that it allows software developers to find hidden faults caused by design flaws which are usually hard to detect. 
Generally, we found that software developers are able to formulate domain-specific constraints in OOASP after a short training. However, existence of ASP development environments supporting debugging and testing of ASP programs would greatly simplify this process.

\subsection{Completion of an instantiation}

The completion task is often applied in situations when a software developer needs to generate a test case for a production system that outputs an invalid instantiation. 
Thus, the completion task allows a developer to detect two types of problems: (i) invalid partial instantiation and (ii) incomplete partial instantiation. 
In the last case, the partial instantiation returned by a configurator can be extended to a valid one by adding missing objects and/or relations between them.
This indicates that the already implemented production system works correctly, at least for the given input, but it is incomplete. The developer can export the obtained solution and use it as a test case during subsequent implementation of the system. 
If the problem of the first type is found, then we have to differentiate between two causes of this problem: (a) the model designed in the CSL Studio is inconsistent; and (b) the system returned a partial instantiation that is faulty, i.e.\ cannot be extended to a valid solution. The first cause can easily be detected by running a completion task with an empty instantiation. If the model is consistent, then manually coded additional constraints of the production system are faulty and the software developer has to correct them.

In order to execute the completion task the CSL Studio exports an instantiation obtained by an object-oriented system to OOASP-DDL. Then, this instantiation together with a corresponding meta-program is provided to an ASP solver. The returned answer sets are visualized by the system to the software developer. 
If needed, the developer can export the found complete instantiation to an instantiation of the object-oriented system. This translation is straight-forward due to the one-to-one correspondence between instances on the OOASP-level and the object-oriented system. 

\begin{figure}[h]
\centerline{\includegraphics[width=3in]{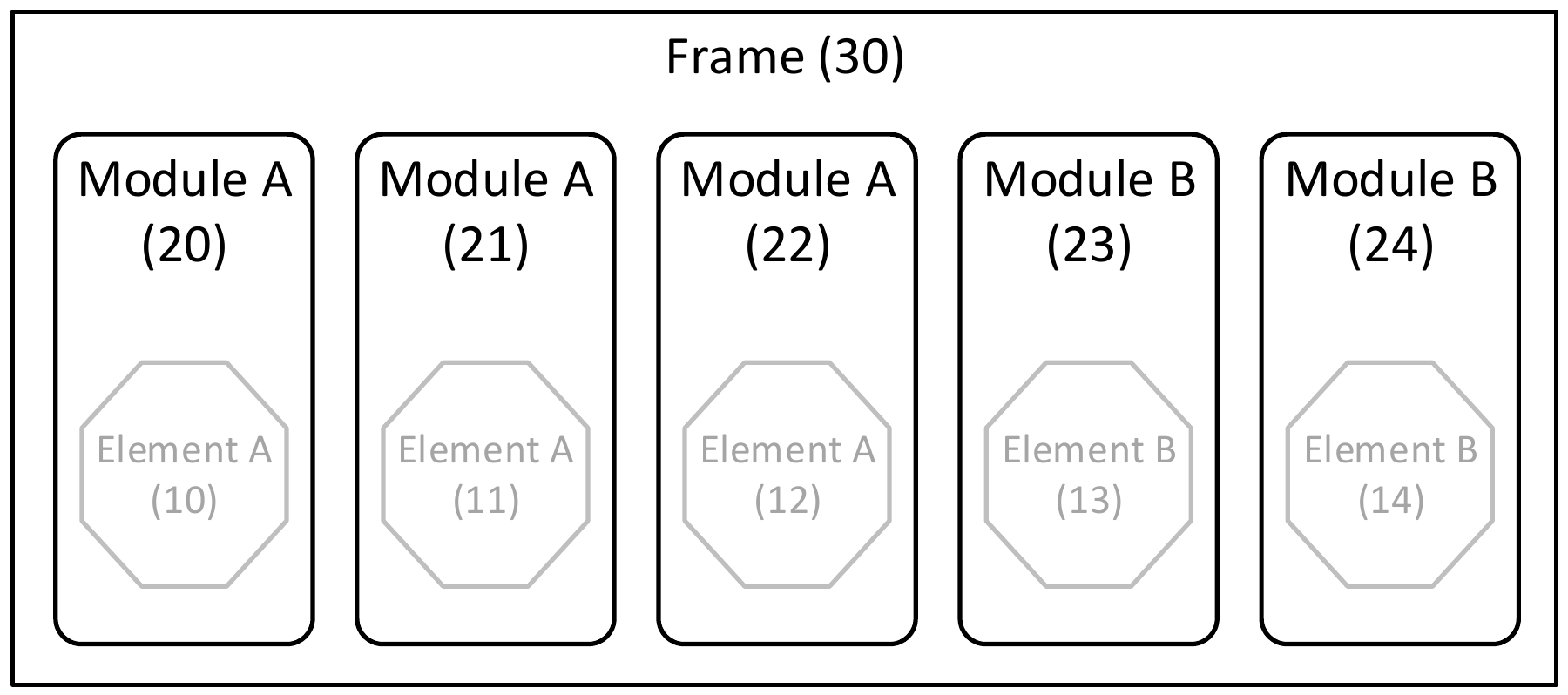}}
\caption{Complete instantiation for the Modules example. The objects existing in the input instantiation are shown in gray.}
\label{modulesconfig1}
\end{figure}

Consider an example in which a partially implemented configuration system returns an instantiation containing three instances of
\texttt{ElementA} and two instances of \texttt{ElementB}. 
\begin{lstlisting}
% Partial configuration 
ooasp_instantiation("v1","c1").
ooasp_isa("c3","ElementA",10). ooasp_isa("c3","ElementA",11).
ooasp_isa("c3","ElementA",12).
ooasp_isa("c3","ElementB",13). ooasp_isa("c3","ElementB",14).
\end{lstlisting}
In this case the completion task returns a solution visualized in Fig.~\ref{modulesconfig1}. This solution comprises the existing objects with identifiers \texttt{10} -- \texttt{14} as well as the new objects corresponding to a frame with object identifier \texttt{30} and five modules \texttt{20} -- \texttt{24}.

\subsection{Reconciliation of an inconsistent instantiation}

The reconciliation task deals with restoring consistency of an inconsistent (partial) instantiation given as an input. The problem arises in three scenarios: (1) the validation task finds an instantiation inconsistent; (2) the completion task detects that a model is consistent, but the given partial  instantiation cannot be extended; and (3) the model is changed due to new requirements to a configurable product. In order to restore the consistency of an instantiation the  reconciliation task comprises two meta-programs. One meta-program converts the input OOASP-DDL program into a reified form. This program comprises rules of the form:
\begin{equation*}
fact(ooasp(\vec{t})) \imp ooasp(\vec{t}).
\end{equation*}
where $ooasp(\vec{t})$ stands for one of the OOASP-DDL atoms listed in Table~\ref{tab:inst}. The second meta-program takes the output of the first one as an input and outputs a consistent instantiation as well as a set of changes applied to obtain it. The set of changes is obtained by the application of deletion/reuse rules of the form:
\begin{align*}
1\{reuse(ooasp(\vec{t})),delete(ooasp(\vec{t}))\}1 & \imp fact(ooasp(\vec{t})). \\
ooasp(\vec{t}) & \imp reuse(ooasp(\vec{t})).
\end{align*}
A preferred solution can be found if a developer provides costs for reuse/delete actions performed by the reconciliation task.

\begin{figure}[h]
\centerline{\includegraphics[width=3in]{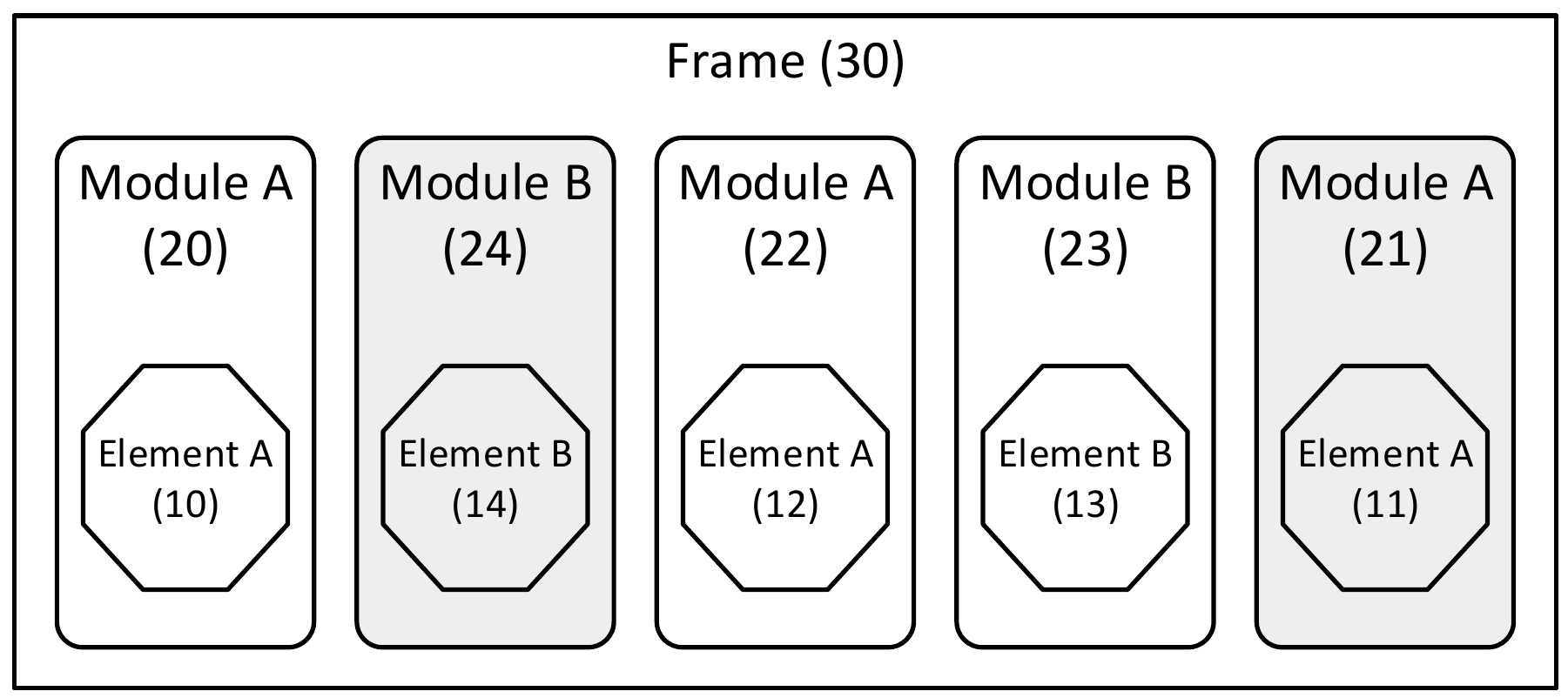}}
\caption{Reconciled configuration for the Modules example}
\label{modulesconfigreconciled}
\end{figure}

For example, suppose that the developer created a configuration system that does not implement a domain-specific constraint preventing overheating of the system. Namely, this constraint avoids overheating by disallowing putting two modules of type \texttt{ModuleA} next to each other.
\begin{lstlisting}
%  do not put 2 modules of type ModuleA next to each other
ooasp_cv(IID,moduleANextToOther(M1,M2,P1,P2)):- 
  ooasp_instantiation("v2",IID),
  ooasp_associated(IID,"Frame_modules",F,M1),
  ooasp_associated(IID,"Frame_modules",F,M2),
  ooasp_attribute_value(IID,"position",M1,P1),
  ooasp_attribute_value(IID,"position",M2,P2),
  M1!=M2,
  ooasp_isa(IID,"ModuleA",M1),
  ooasp_isa(IID,"ModuleA",M2),
  P2=P1+1.
\end{lstlisting}
Due to the added constraint, the instantiation in Fig.~\ref{modulesconfig1} is no longer valid. The reconciliation task finds a required change by modifying the positions of modules with identifiers \texttt{21} and \texttt{24}. The result of the reconciliation can be presented to a developer by OOASP framework as shown in Fig.~\ref{modulesconfigreconciled}.

\section{Conclusions}\label{sec:conc}

This paper demonstrates OOASP which integrates ASP into the object-oriented software development process using an industrial product configurator as an evaluation example. Our preliminary results are very encouraging and open a number of new directions for a tighter integration of object-oriented programming and ASP. 
Thus, our experiments with OOASP showed that checking constraints with respect to a given object-oriented model can be done efficiently by modern ASP solvers. 
However, execution of the reconciliation task still remains a challenge for large-scale instantiations~\cite{Friedrich2011a}. 
It appears that the main obstacle for the approach based on ASP meta-programming is the explosion of grounding.
In addition, the completion of large-scale instantiations indicated that a computation time for a solution can be improved by the application of domain-specific heuristics. The latter are often hard to implement for software developers, since they do not have enough experience in ASP. In our future work we are going to investigate these questions in more details.

\end{document}